\newcommand{\CLASSINPUTbottomtextmargin}{1in}
\documentclass[letterpaper,conference]{IEEEtran}
\IEEEoverridecommandlockouts    
\usepackage{multirow}
\usepackage{lipsum}
\usepackage{amsmath}
\usepackage{amssymb}
\usepackage[ruled,vlined]{algorithm2e}
\usepackage{graphicx}
\graphicspath{{./Figures/}}
\usepackage{threeparttable}
\usepackage{longtable}
\usepackage{tcolorbox}
\usepackage{tabularx}
\title{\LARGE \bf A Comparative Study of Graph Neural Network Layer Selection for Interaction Modelling in Driving Trajectory Prediction}

 \author{
 	\parbox{\textwidth}{%
 		\centering
 		George Daoud$^{1,2}$, Mohamed El-Darieby$^{1}$%
 	}%
 	\thanks{$^{1}$Ontario Tech University, Oshawa, ON, Canada
 		{\tt\small George.Daoud@OntarioTechU.ca, Mohamed.El-Darieby@OntarioTechU.ca}}%
 	\thanks{$^{2}$Assiut University, Assiut, Egypt
 		}%
 }

\begin{document}
	
	\maketitle
	\thispagestyle{empty}
	\pagestyle{empty}
	
	\begin{abstract}
Autonomous driving systems rely on precise trajectory prediction to plan safe and efficient movement. Graph Neural Networks (GNNs) have become a promising approach for modelling spatiotemporal interactions among road agents. However, designing GNN architectures for trajectory prediction remains non-standardized, with little guidance on which graph layers effectively capture spatial interactions and temporal dynamics. This paper offers a detailed comparative study of 19 graph layer types, focusing on their spatial and temporal processing capabilities to discover the most effective architectures for trajectory prediction. Within the explored hyperparameter setting, we highlight five standout layer combinations, with ARMA, Chebyshev, and topology-aware layers consistently performing better than others. Beyond performance metrics, our findings yield practical design principles: sum-based aggregation is more effective than mean-based methods, multi-head attention mechanisms enable richer interactions, and assigning different weights to different hop distances significantly improves prediction accuracy. These findings offer useful guidance for designing more interpretable and effective trajectory prediction models.

	\end{abstract}
	
	\section{Introduction}
	\label{sec:introduction}
Predicting the future paths of road agents, such as vehicles, pedestrians, and cyclists, is now a crucial part of autonomous driving systems. This process works between the perception and planning modules. By modelling interactions among nearby agents and the ego vehicle, trajectory prediction allows for safe and efficient short-term planning \cite{Huang2022}. This prediction is beneficial not only for autonomous driving but also for traffic safety assessment and adaptive control in Intelligent Transportation Systems (ITS), especially in complex environments such as highways and roundabouts \cite{Zhao2020}.

Current trajectory prediction methods can be grouped into physics-based and machine-learning-based approaches. Physics-based methods use physical and probabilistic models, while learning-based methods derive motion patterns directly from data. Although learning-based approaches usually offer higher accuracy, they often struggle to capture the semi-structured, dynamic nature of driving scenes and the complex interactions between agents, which involve changes over time, varying road geometries, and varying numbers of agents \cite{Daoud2023}.

A common solution is to represent driving scenes as sequences of semantic bird's-eye-view (BEV) images, where interactions are reflected in pixel locations and values. While BEV representations provide fixed-size inputs, they need discretization, increase input dimensionality, and come with higher computational costs \cite{madjid2025trajectorypredictionautonomousdriving}.

Graph-based representations provide a more organized alternative by modelling agents as nodes and their interactions as edges, either across individual time steps or within a single spatiotemporal graph. Road network information can be incorporated through node features \cite{Daoud2023_2} or by explicitly modelling roads and lanes as graphs \cite{gao2020vectornetencodinghdmaps}.

Graph-based trajectory prediction models utilize graph neural networks (GNNs) to capture spatial and temporal interactions while keeping input dimensionality low and consistent. Despite promising results, the effectiveness of different GNN layer types for modelling interactions remains unclear. To fill this gap, this paper presents a detailed comparative study of graph neural layers for spatiotemporal trajectory prediction and expands on the architecture of Daoud et al. \cite{Daoud2023_2} to enable longer prediction horizons with shorter observation windows.

This work makes three contributions: (1) an evaluation of 19 graph convolutional layer types for spatiotemporal trajectory prediction, filling a critical gap in GNN architecture design; (2) identification of five superior layer combinations that outperform prior work on roundabout scenarios; and (3) design principles that practitioners can apply to GNN-based trajectory prediction systems, such as the superiority of sum-based aggregation and the importance of hop-specific weight matrices.
	\section{Related Works}
	\label{sec:RelatedWorks}

Existing trajectory prediction methods can be divided into physics-based and machine-learning-based approaches. Physics-based models apply physical laws and probabilistic techniques to understand interactions and predict future movement. For instance, Kalman Filters have been used with kinematic models like constant turn rate and acceleration (CTRA) to address uncertainty \cite{Xie2018}. Additionally, comfort constraints can be added to cost functions to create smooth trajectories \cite{Sorstedt2011}.

Machine-learning approaches learn driving patterns from data. They can be further categorized by how they represent input into bird's-eye-view (BEV) and graph-based models. BEV-based methods rasterize scenes into semantic images and employ deep neural networks such as CNNs \cite{Mandal2020}, VAEs \cite{Xu2023} and conditional VAEs \cite{Zhong2022}, or LSTM-based encoder-decoder models with attention \cite{Messaoud2021} or social pooling \cite{Messaoud2019} to predict future trajectories, mainly for the ego vehicle. Attention mechanisms are frequently used to understand interactions between cars and road elements \cite{Zhang2025}.

Graph-based models depict driving scenes as graphs. They can be structured as sequences of spatial graphs with temporal propagation, as a single spatiotemporal graph, or as heterogeneous graphs that feature different types of spatial and temporal edges. Sequence-based methods handle spatial interactions using graph convolutional layers (GCLs) and capture temporal dynamics with sequence-to-sequence architectures like Transformers \cite{Zhang2022} or GRUs \cite{Zhang2023}. Single-graph formulations embed temporal data into node features and typically use GAT layers, which have been shown to perform better than GCNs \cite{Diehl2019}. Heterogeneous graph models clearly differentiate spatial and temporal interactions by using different GCLs, such as GAT \cite{veličković2018graphattentionnetworks} for spatial relationships and GCN \cite{kipf2017semisupervisedclassificationgraphconvolutional} for temporal dynamics \cite{Daoud2023_2}.

Hybrid methods mix physics-based and learning-based models. For example, they may integrate shock-wave physics \cite{Yao2023}, combine kinematic models with learned predictors \cite{Kim2021}, or merge physical model outcomes with recurrent networks \cite{Li2023}. Additional interaction modelling techniques, such as game-theoretic approaches, inverse reinforcement learning, and drift-diffusion models, are reviewed by Wang et al. \cite{Wenshuo_2208.07541}.

This paper focuses on heterogeneous graph-based models, which blend the precision of learning-based methods with the effectiveness of graph representations. Building on Daoud et al. \cite{Daoud2023_2}, we present a modified architecture and conduct a comparative study to identify the most effective graph convolutional layers for predicting vehicle trajectories.

	\section{Architecture Design and Layer Selection}
	\label{sec:ArchitectureDesignandLayerSelection}

The proposed architecture builds on the architecture proposed by Daoud et al. \cite{Daoud2023_2}, aiming to improve prediction accuracy and extend the prediction horizon. We also evaluate 19 graph convolutional layers (GCLs) to find the most effective configurations.

Unlike the original approach, which rotates the map to match the target vehicle's heading, we keep the map centered without rotation. Since map segments are already globally aligned, this simplification reduces computation while maintaining route feasibility. As a result, the model uses a shorter 1-second observation window and a longer 5-second prediction horizon.

\begin{figure*}[ht] \centering \includegraphics[width=\linewidth, height=4cm]{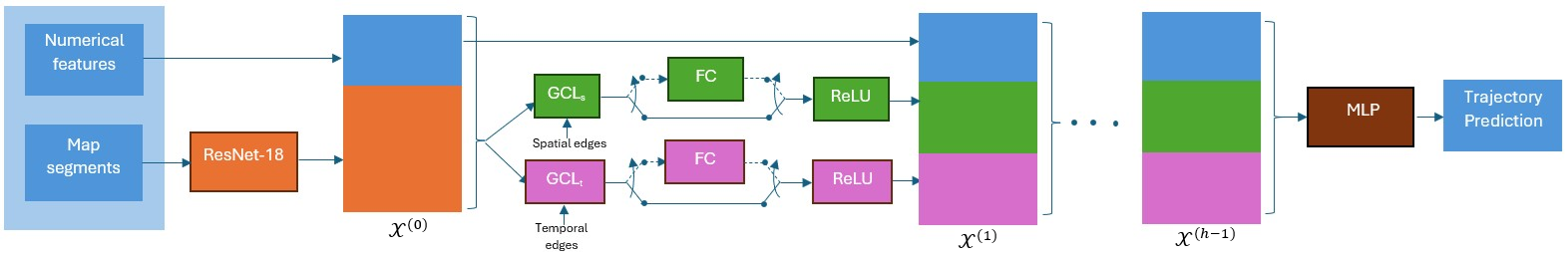} \caption{The general view of the proposed architecture} \label{fig:arch} \end{figure*}

Figure~\ref{fig:arch} shows the overall architecture. Map data is processed independently using a ResNet-18 to generate a compact embedding. This embedding is then combined with numerical agent features to create the initial node representation. Node embeddings are updated over $h$ iterations, which corresponds to the number of historical frames, using paired spatial and temporal GCLs to model interactions. Skip connections are included to reduce over-smoothing and improve expressiveness. A final MLP produces the predicted trajectory. The map embedding is 200-dimensional, while spatial and temporal embeddings are 100-dimensional. The driving scenes are sampled at 5 Hz, using five historical frames to predict 25 future steps. Four GCL layers are used, and the MLP outputs a 50-dimensional vector per node. When specific GCLs need it, an optional fully connected layer projects inputs to a compatible dimension (100).

To choose suitable GCLs, we assess 18 variants across spatial (${GCL}_s$) and temporal (${GCL}_t$) components. Starting from a GCN \cite{kipf2017semisupervisedclassificationgraphconvolutional} baseline, layers are replaced one at a time and in combination. Table~\ref{tab:gcls} summarizes the evaluated GCLs and their parameters. The GCLs are divided into six categories:

\begin{table}
  \caption{The list of graph convolutional layers and their parameters}
  \begin{tabularx}{0.5 \textwidth}{c|c}
    \hline
    Graph convolutional layer&Parameters $\ast$ \\
    \hline
    Graph Convolutional Network (GCN) \cite{kipf2017semisupervisedclassificationgraphconvolutional}  & \\
    SAGE \cite{hamilton2018inductiverepresentationlearninglarge}&  \\
    Higher-order Graph Networks (HoGraph) \cite{morris2021weisfeilerlemanneuralhigherorder}&\\
    \hline
    Graph attentional layer (AGNN) \cite{thekumparampil2018attentionbasedgraphneuralnetwork}&needs an extra FC\\
    Frequency Adaptive Convolution (FA) \cite{bo2021lowfrequencyinformationgraphconvolutional}&$\epsilon=0.1$\\
    Graph Attention Network (GAT) \cite{veličković2018graphattentionnetworks} &\\
    Local Extremum Network (LEConv) \cite{ranjan2020asapadaptivestructureaware}&\\
    \hline
    Efficient Graph Convolution (EGC) \cite{tailor2022needanisotropicgraphneural}& $H=4$ , $B=4$\\
    Transformer \cite{shi2021maskedlabelpredictionunified}&$H=4$\\
    SuperGAT \cite{kim2022friendlyneighborhoodgraphattention} & $H=4$ \\
    
    \hline 
    Simplifying Graph Convolutional (SGC) \cite{wu2019simplifyinggraphconvolutionalnetworks} & K=3 \\
     Simple Spectral Convolutional (S$^{2}$GC) \cite{zhu2021simple}&$\alpha=0.5$ , $K=3$ \\
    MixHop  \cite{abuelhaija2019mixhophigherordergraphconvolutional}&K=3\\
    Topology adaptive convolutional (TAGCN) \cite{du2018topologyadaptivegraphconvolutional}& K=3 \\
    Molecular Fingerprints (MF) \cite{duvenaud2015convolutionalnetworksgraphslearning}&\\
     
     \hline
    Gated Graph Convolution (GRU) \cite{li2017gatedgraphsequenceneural} &needs an extra FC\\
     Residual Gated Convolutional (ResGRU) \cite{bresson2018residualgatedgraphconvnets}&\\
     
     \hline
    ARMA \cite{Bianchi_2021} & $T_L=1, K_s=1 $\\
    Chebyshev Spectral Graph Convolutional \cite{defferrard2017convolutionalneuralnetworksgraphs} & $k_c=3$ \\
     \hline
\end{tabularx}
    \begin{tablenotes}
      \footnotesize
      \item $\ast$ $H$: the number of heads, $K$: the number of hops, $K_s$ and $T_L$: the number of stacks and layers for ARMA filter, $K_c$: Chebyshev filter length, $B$: the number of bases
    \end{tablenotes}
\label{tab:gcls}
\end{table}

\begin{enumerate}
    \item Traditional Graph Convolutions, that include GCN, GraphSAGE\cite{hamilton2018inductiverepresentationlearninglarge}, and Higher-order graph (HoGraph) \cite{morris2021weisfeilerlemanneuralhigherorder}. These methods aggregate neighbour information by summation or averaging. They use self-loops to combine features from nodes and their neighbours.
    
    \item Single-Head Attention-Based layers, which assign attention weights to neighbours based on how relevant they are. Examples include AGNN \cite{thekumparampil2018attentionbasedgraphneuralnetwork}, Frequency
Adaptation Graph (FA) \cite{bo2021lowfrequencyinformationgraphconvolutional}, GAT \cite{veličković2018graphattentionnetworks}, and LEConv \cite{ranjan2020asapadaptivestructureaware}. They use cosine similarity or learned projections.
    
    \item Multi-Head Attention-Based layers, that extend attention mechanisms with multiple heads to capture different interactions. This group includes EGC \cite{tailor2022needanisotropicgraphneural}, Transformer-based layers \cite{shi2021maskedlabelpredictionunified}, and SuperGAT \cite{kim2022friendlyneighborhoodgraphattention}. All of them use four attention heads.
    
    \item Topology-Based layers, that take advantage of the multi-hop structure of graphs. For example, SGC \cite{wu2019simplifyinggraphconvolutionalnetworks} and S$^{2}$GC \cite{zhu2021simple} share weights across hops. MixHop \cite{abuelhaija2019mixhophigherordergraphconvolutional} and TAGCN \cite{du2018topologyadaptivegraphconvolutional} use hop-specific weights. Molecular Fingerprints (MF)  \cite{duvenaud2015convolutionalnetworksgraphslearning} adjusts the weights based on the node's degree.
    
    \item Recurrent-Based layers, which use recurrent units like GRUs. They update node embeddings by treating neighbour information as sequential input.
    
    \item Specialized Graph filters, that include ARMA \cite{Bianchi_2021} and Chebyshev \cite{defferrard2017convolutionalneuralnetworksgraphs}  convolutions. They use spectral filtering to capture long-range dependencies.
\end{enumerate}

	\section{Experiments}
	\label{sec:Experiments}

The proposed model is evaluated using the RounD dataset \cite{rounDdataset}, which was collected in Germany utilizing a drone hovering over multiple roundabouts. It contains detailed trajectory recordings of various road users, including bicycles, motorcycles, cars, trailers, trucks, vans, and buses.

In this study, experiments are conducted on the third scenario of the dataset, which captures traffic activity at a roundabout. Graph construction follows three rules: (1) spatial edges connect agents in the same frame within 30m Euclidean distance; (2) temporal edges link consecutive agent instances across frames; (3) nodes lacking a complete 5-frame history or 25-frame prediction horizon are masked from the prediction. A summary of the resulting graph structure is provided in Table~\ref{tab:graph}.

\begin{table}
    \centering
    \caption{Overview of graph structure and preprocessing rules }
    \begin{tabular}{c|c}
        \hline
         Metric&Value\\
         \hline
Original sampling& 25 FPS, downsampled to 5 FPS\\
History / Horizon & 5 frames (1s) / 25 frames (5s)\\
Graph size&5.4K frames, 54.4K nodes\\
Edges&203.9K spatial, 53.7K temporal\\
Valid nodes&34K (62.5\%)\\
Train/Val/Test frames&70\% / 20\% / 10\% \\
Batch size & 1 frame \\
Numerical input data & Dimension, heading \& Type \\
Output data & Predicted Trajectory
    \end{tabular}
    
    \label{tab:graph}
\end{table}

During training, the model parameters are optimized using the Adam optimizer. Training is performed for 60 epochs, with the learning rate reduced by a factor of 10 after 30 epochs and again after an additional 20 epochs. The initial learning rate is set to $10^{-3}$ for all models except EGC, which uses $10^{-4}$. Mean Squared Error (MSE) is used as the training loss function. All experiments are conducted on a workstation equipped with 64 GB of RAM and an NVIDIA GeForce RTX 4090 GPU. On average, training a single model requires approximately 10 hours.

At the end of each epoch, the model is evaluated on the validation set using the Average Displacement Error (ADE), defined in Equation~(\ref{eq:ADE}). The best-performing model is saved and restored before each learning rate reduction to ensure optimal convergence. After training is complete, a final evaluation is performed on the test set using both the ADE and the Final Displacement Error (FDE), defined in Equation~(\ref{eq:FDE}).

\begin{equation}
    ADE(\Delta T_f) = \frac{1}{\Delta T_f} \sum_{\Delta t_f=1}^{\Delta t_f=\Delta T_f}{FDE(\Delta t_f)}
    \label{eq:ADE}
\end{equation}

\begin{equation}
    FDE(\Delta t_f) = \frac{1}{|\mathcal{N}|} \sum_{n\in \mathcal{N}}{\left\| \begin{bmatrix} \Delta \hat{x}_{n}(\Delta t_f) - \Delta x_{n}(\Delta t_f) \\
    \Delta \hat{y}_{n}(\Delta t_f) - \Delta y_{n}(\Delta t_f) \\
    \end{bmatrix} \right\|_{2}}
    \label{eq:FDE}
\end{equation}

Here, $\Delta t_f$ denotes the prediction time step, and $\Delta T_f$ represents the full prediction horizon. The set $\mathcal{N}$ includes all nodes with complete history and future. $\Delta x_{n}$ and $\Delta y_{n}$ denote the ground-truth future displacement of node $n$, while $\hat{x}_{n}$ and $\hat{y}_{n}$ represent the corresponding predicted displacements.
\section{Results and Discussion}
\label{sec:ResultsandDiscussion}
For consistency with prior work, Graph Convolutional Networks (GCNs) were used as the baseline for both spatial and temporal layers ($\text{GCL}_{s}$ and $\text{GCL}_{t}$). Following standard ablation practices, each layer was first replaced independently and then jointly with alternative graph layers. Performance was evaluated using Average Displacement Error (ADE) and Final Displacement Error (FDE) over prediction horizons at 3 and 5 seconds. For comparison, we also report results from three studies using the same dataset: a BEV-based CNN model \cite{Nikhil2019} (as reported by \cite{Steiner2024}), a deep GNN approach \cite{Daoud2023}, and a hybrid GNN–CNN–Transformer model \cite{Steiner2024}. The latter reports $minADE$ and $minFDE$ due to its multimodal predictions, making comparison more challenging since our model produces a single trajectory. Nevertheless, their results are included in Table~\ref{tab:results0}. For clarity, superior values in the following tables will be highlighted.

\begin{table}
\centering
  \caption{Trajectory Prediction results from literature}

  \begin{tabular}{cc|cc|cc}
  \hline
    \multicolumn{2}{c|}{Model}&\multicolumn{2}{c|}{$ADE$}&  \multicolumn{2}{c}{$FDE$}\\
    &&@3s&@5s&@3s&@5s\\
    \hline
    \multicolumn{2}{c|}{CNN+LSTM \cite{Nikhil2019}}  & 1.46 & 3.57 & 4.30 & 10.29 \\    
    \multicolumn{2}{c|}{Extended DGNN \cite{Daoud2023}} & ---& 1.69 & --- & ---     \\
    \multicolumn{2}{c|}{MAP-FORMER \cite{Steiner2024} $\ast$}  &0.52 &1.20& 1.38 & 3.22 \\
\hline
    \end{tabular}
    \begin{tablenotes}
      \footnotesize
      \item $\ast$ uses the $minADE$ and $minFDE$ instead of $ADE$ and $FDE$.
    \end{tablenotes}
\label{tab:results0}
\end{table}
Table~\ref{tab:results1} summarizes results for traditional GCLs. Both GraphSAGE and HoGraph outperform the GCN baseline, with HoGraph achieving the best results across spatial and temporal layers. In particular, it surpasses prior work in terms of $FDE$ at 5 seconds. These findings suggest that using separate weight matrices for nodes and their neighbours improves performance. In addition, sum-based aggregation consistently outperforms mean aggregation in this setting.

\begin{table}
\centering
  \caption{Trajectory Prediction results for traditional GCLs}
  \begin{tabular}{cc|cc|cc}
    \hline
    \multicolumn{2}{c|}{Model}&\multicolumn{2}{c|}{$ADE$}&  \multicolumn{2}{c}{$FDE$}\\
    $\text{GCL}_{s}$ & $\text{GCL}_{t}$ &@3s&@5s&@3s&@5s\\
    \hline
    GCN (base) &GCN (base) &	1.08 & 1.89&	2.06  &	4.14 \\
        \hline
    GCN (base)  & SAGE       &0.84&1.58&1.70&3.72\\
    SAGE        &GCN (base)  &0.99&1.70&1.84&3.73\\
    SAGE        & SAGE       &0.79&1.44&1.55&3.30\\
    \hline
    GCN (base)  &HoGraph    &0.92&1.69&1.82&3.85\\
    HoGraph     &GCN (base) &0.87&1.47&1.59& \colorbox{lightgray}{3.21}\\
    HoGraph     &HoGraph    &0.77&1.34&1.44&\colorbox{lightgray}{3.06}\\
    \hline
    \end{tabular}
\label{tab:results1}
\end{table}

Table~\ref{tab:results2} reports the performance of single-head attention models. AGNN and FA generally underperform the baseline, whereas GAT consistently improves prediction accuracy. LEConv achieves the best performance in this group and outperforms prior work in terms of $FDE$ at 5 seconds. These results indicate that transforming node embeddings before computing attention (as in GAT) is more effective than using cosine similarity or raw features. Avoiding projection altogether, as in LEConv, yields further gains.

\begin{table}
\centering
  \caption{Trajectory Prediction results for single-head attention-based GCLs}
  \begin{tabular}{cc|cc|cc}
    \hline
    \multicolumn{2}{c|}{Model}&\multicolumn{2}{c|}{$ADE$}&  \multicolumn{2}{c}{$FDE$}\\
    $\text{GCL}_{s}$ & $\text{GCL}_{t}$ &@3s&@5s&@3s&@5s\\
    \hline
   GCN (base)  &AGNN       &1.11&1.93&2.11&4.25\\
    AGNN        &GCN (base) &1.20&2.03&2.21&4.35\\
    AGNN        &AGNN       &1.11&1.94&2.12&4.28\\
    \hline
    GCN (base)  &FA         &1.20&2.09&2.28&4.59\\
    FA          &GCN (base) &1.02&1.77&1.93&3.89\\
    FA          &FA         &1.20&2.03&2.22&4.31\\
    \hline
    GCN (base) &GAT        &0.94&1.62&1.76&3.57 \\
    GAT        &GCN (base) &0.95&1.64&1.78&3.61 \\
    GAT	       &GAT        & 0.91&1.58&1.70&3.52 \\
    \hline
    GCN (base)  &LEConv         &0.88&1.56&1.66&3.55\\
    LEConv          &GCN (base) &0.77&1.33&1.42&\colorbox{lightgray}{3.03}\\
    LEConv          &LEConv         &0.85&1.45&1.57&\colorbox{lightgray}{3.19}\\
    \hline
    \end{tabular}
\label{tab:results2}
\end{table}

Results for multi-head attention models are shown in Table~\ref{tab:results3}. Transformer and SuperGAT significantly outperform the baseline, whereas EGC fails to improve performance due to its reliance on basis decomposition rather than head-specific transformations. SuperGAT achieves the strongest results in this category, thanks to its expressive attention formulation, which better captures inter-agent interactions.

\begin{table}
\centering
  \caption{Trajectory Prediction results for multi-head attention-based GCLs}
  \begin{tabular}{cc|cc|cc}
    \hline
    \multicolumn{2}{c|}{Model}&\multicolumn{2}{c|}{$ADE$}&  \multicolumn{2}{c}{$FDE$}\\
    $\text{GCL}_{s}$ & $\text{GCL}_{t}$ &@3s&@5s&@3s&@5s\\
    \hline

    GCN (base)  &EGC        &1.23&2.22&2.42&4.94\\
    EGC         &GCN (base) &1.37&2.46&2.72&5.40\\
    EGC         &EGC        &1.49&2.62&2.87&5.68\\
    \hline
     GCN (base)  &Transformer&0.97&1.70&1.83&3.79\\
    Transformer &GCN (base) &0.75&1.32&1.40&\colorbox{lightgray}{3.00}\\
    Transformer &Transformer&0.70&1.26&\colorbox{lightgray}{1.34}&\colorbox{lightgray}{2.91}\\
        \hline
    GCN (base)  &SuperGAT    &0.92&1.70&1.84&3.92\\
    SuperGAT    &GCN (base)  &0.76&1.30&1.40&\colorbox{lightgray}{2.87}\\
    SuperGAT    &SuperGAT    &0.78&1.40&1.51&\colorbox{lightgray}{3.19}\\
    \hline
    \end{tabular}
\label{tab:results3}
\end{table}

Table~\ref{tab:results4} presents results for topology-aware GCLs. SGC and S$^{2}$GC perform poorly due to shared weights across hops. In contrast, MixHop and TAGCN perform better by assigning distinct weights to different hop distances. TAGCN achieves the strongest performance, especially for long-horizon prediction, by using summation rather than concatenation. MF also performs competitively by adapting weights based on node degree, though it slightly underperforms TAGCN at 3 seconds.

\begin{table}
\centering
  \caption{Trajectory Prediction results for topology-based GCLs}
  \begin{tabular}{cc|cc|cc}
    \hline
    \multicolumn{2}{c|}{Model}&\multicolumn{2}{c|}{$ADE$}&  \multicolumn{2}{c}{$FDE$}\\
    $\text{GCL}_{s}$ & $\text{GCL}_{t}$ &@3s&@5s&@3s&@5s\\
    \hline
    
    GCN (base)  &SGC         &1.12&1.94&2.12&4.25\\
    SGC          &GCN (base) &1.09&1.88&2.04&4.11\\
    SGC          &SGC         &1.00&1.72&1.88&3.77\\
    \hline
    GCN (base)  &S$^{2}$GC        &1.01&1.73&1.88&3.77\\
    S$^{2}$GC         &GCN (base) &1.01&1.74&1.89&3.81\\
    S$^{2}$GC         &S$^{2}$GC        &1.10&1.94&2.09&4.32\\
    \hline
    GCN (base)  &MixHop     &1.07&1.87&2.02&4.16\\
    MixHop      &GCN (base) &0.97&1.68&1.83&3.67\\
    MixHop      &MixHop     &1.11&1.96&2.14&4.30\\
    \hline
    GCN (base)  &TAGCN        &0.83&1.50&1.60&3.45\\
    TAGCN         &GCN (base) &0.85&1.43&1.53&\colorbox{lightgray}{3.15}\\
    TAGCN	        &TAGCN&0.66&\colorbox{lightgray}{1.19}&\colorbox{lightgray}{1.25}&\colorbox{lightgray}{2.80}\\
    \hline
    GCN (base)  &MF         &0.84&1.48&1.58&3.35\\
    MF          &GCN (base) &0.67&\colorbox{lightgray}{1.15}&\colorbox{lightgray}{1.25}&\colorbox{lightgray}{2.54}\\
    MF          &MF         &0.81&1.38&1.50&\colorbox{lightgray}{3.03}\\
  \hline
\end{tabular}
\label{tab:results4}
\end{table}

Table~\ref{tab:results5} summarizes the performance of GRU-based models. Both GRU and ResGRU outperform prior work when applied to the spatial layer. ResGRU consistently achieves better results due to its combination of recurrent modelling and attention mechanisms.

\begin{table}
      \caption{Trajectory Prediction results for GRU-based GCLs}
  \begin{tabular}{cc|cc|cc}
    \hline
    \multicolumn{2}{c|}{Model}&\multicolumn{2}{c|}{$ADE$}&  \multicolumn{2}{c}{$FDE$}\\
    $\text{GCL}_{s}$ & $\text{GCL}_{t}$ &@3s&@5s&@3s&@5s\\
    \hline
        GCN (base)  &GRU        &1.03&1.79&1.92&3.95\\
    GRU         &GCN (base) &0.81&1.40&1.50&\colorbox{lightgray}{3.15}\\
    GRU         &GRU        &0.91&1.67&1.81&3.84\\
    \hline
    GCN (base)  &ResGRU     &0.97&1.83&1.99&4.22\\
    ResGRU      &GCN (base) &0.73&1.24&\colorbox{lightgray}{1.32}&\colorbox{lightgray}{2.74}\\
    ResGRU      &ResGRU     &0.75&1.31&1.40&\colorbox{lightgray}{2.95}\\
\hline
\end{tabular}
\label{tab:results5}
\end{table}

Results for spectral and filter-based layers are shown in Table~\ref{tab:results6}. ARMA and Chebyshev convolutions achieve strong performance and outperform prior methods across most metrics. When applied to both spatial and temporal layers, these models yield the best overall results, particularly for longer prediction horizons.

\begin{table}
      \caption{Trajectory Prediction results for special types of GCLs}
  \begin{tabular}{cc|cc|cc}
    \hline
    \multicolumn{2}{c|}{Model}&\multicolumn{2}{c|}{$ADE$}&  \multicolumn{2}{c}{$FDE$}\\
    $\text{GCL}_{s}$ & $\text{GCL}_{t}$ &@3s&@5s&@3s&@5s\\
    \hline
    GCN (base)  &ARMA       &0.85&1.58&1.71&3.68\\
    ARMA        &GCN (base) &0.82&1.41&1.51&\colorbox{lightgray}{3.14}\\
    ARMA        &ARMA       &0.67&\colorbox{lightgray}{1.19}&\colorbox{lightgray}{1.28}&\colorbox{lightgray}{2.74}\\
    \hline
    GCN (base)  &Chebyshev  &0.96&1.66&1.79&3.67\\
    Chebyshev   &GCN (base) &0.72&\colorbox{lightgray}{1.20}&\colorbox{lightgray}{1.30}&\colorbox{lightgray}{2.63}\\
    Chebyshev   &Chebyshev  &0.65&\colorbox{lightgray}{1.10}&\colorbox{lightgray}{1.19}&\colorbox{lightgray}{2.45}\\
    
  \hline
\end{tabular}
\label{tab:results6}
\end{table}

Table~\ref{tab:con} summarizes the best-performing configurations. The strongest results are obtained using topology-aware and Specialized filter layers. Although the proposed model is unimodal, it outperforms all unimodal baselines and even surpasses some multimodal methods. Performance gains are not additive. For example, replacing GCN with LEConv in the temporal layer improves $ADE@5s$ by 17.4\%, while replacing it in the spatial layer improves it by 29.6\%. However, applying LEConv to both layers yields only a 23.3\% improvement, highlighting the interdependence of spatial and temporal modelling. Overall, spatial modelling has a greater impact on performance, as improvements were observed in 13 of the 18 tested configurations when the spatial GCL was modified.

\begin{table}
\centering
  \caption{Best graph layers for spatiotemporal trajectory prediction}
  \begin{tabular}{c|cc}
    \hline
    Type & $\text{GCL}_{s}$ & $\text{GCL}_{t}$ \\
    \hline
    Topology-based & TAGCN & TAGCN \\
    Topology-based & MF & GCN \\
    \hline
    Specialized Filters & ARMA & ARMA \\
    Specialized Filters & Chebyshev & GCN \\
    Specialized Filters & Chebyshev & Chebyshev \\
    \hline
  \end{tabular}
  \label{tab:con}
\end{table}
\section{Conclusion and Future Work}

In this work, we present a graph-based framework for modelling interactions among road agents and predicting road-agent trajectories over a 5s horizon using only 1s of history. The model is evaluated on a roundabout scenario, and an extensive analysis is conducted to study the effect of different graph convolutional layer designs. By evaluating combinations of 18 graph layers, the following conclusions are drawn:

\begin{itemize}
\item Five layer combinations achieve the best performance.
\item Sum-based aggregation is more effective than averaging for traditional GCLs.
\item In single-head attention models, applying a linear transformation before attention improves accuracy.
\item Multi-head attention benefits from using distinct transformations per head.
\item Topology-aware layers perform better when using hop- or degree-specific weights and sum-based aggregation.
\item GRU-based layers require attention mechanisms to achieve competitive results.
\item Spectral filters, particularly Chebyshev convolutions, consistently outperform other layer types.
\item Overall, graph-based models significantly outperform traditional machine learning approaches for trajectory prediction.
\end{itemize}

One limitation of this work is that only 55 graph-layer combinations were evaluated out of a possible $18^2$. While this ablation-style selection highlights the most influential layers, it does not fully explore the design space. Future studies can examine additional combinations, particularly those that performed well when applied exclusively to either the spatial or temporal components.
Another limitation is the use of a single driving scenario type. Although roundabouts present complex interactions and rich traffic dynamics, they do not capture all driving conditions. Evaluating the model on other scenarios is necessary to assess its generalization capability.

This study also assumes that ADE and FDE sufficiently capture prediction quality and that unimodal predictions offer a reasonable trade-off between accuracy and efficiency. In addition, it assumes that performance trends generalize across different types of road users, although further validation is required. Other metrics needed to be investigated as well. Also, Computational cost, as model parameter counts and FLOP estimates, needs to be included in future work.

Finally, a promising direction for future work is the development of hybrid graph layers that combine the most effective properties of the evaluated methods. Integrating strengths from multiple graph convolutional strategies could yield more expressive and robust spatiotemporal models.

	\bibliographystyle{IEEEtran}
	\bibliography{root} 
	
\end{document}